\documentclass[10pt,journal,compsoc]{IEEEtran}

\usepackage[utf8]{inputenc}
\usepackage[T1]{fontenc}
\usepackage{microtype}
\usepackage{url}

\usepackage{amsmath,amsfonts,amssymb}
\usepackage{dsfont}

\usepackage{graphicx}
\usepackage{epsfig}
\usepackage{array}
\usepackage{booktabs}
\usepackage{makecell}
\usepackage{multirow}
\usepackage{tabularx}
\usepackage{colortbl}
\usepackage{diagbox}
\usepackage[table]{xcolor}

\usepackage{pifont}
\usepackage{xspace}
\usepackage{enumitem}
\usepackage[ruled,vlined,linesnumbered]{algorithm2e}
\DontPrintSemicolon
\SetAlgoNlRelativeSize{-1}

\ifCLASSOPTIONcompsoc
  \usepackage[nocompress]{cite}
\else
  \usepackage{cite}
\fi

\usepackage{tcolorbox}
\usepackage{listings}
\tcbuselibrary{breakable,skins,listings}

\usepackage[
  colorlinks=true,
  linkcolor=black,
  citecolor=blue,
  urlcolor=blue,
  bookmarks=true,
  bookmarksnumbered=true
]{hyperref}

\makeatletter
\DeclareRobustCommand\onedot{\futurelet\@let@token\@onedot}
\def\@onedot{\ifx\@let@token.\else.\null\fi\xspace}
\def\eg{\emph{e.g}\onedot}

\def\ie{\emph{i.e}\onedot}

\makeatother

\definecolor{suppheader}{RGB}{47,47,47}
\definecolor{suppgray}{RGB}{246,246,246}
\definecolor{traceheader}{RGB}{196,0,0}
\definecolor{traceback}{RGB}{255,247,247}

\newtcolorbox{resultbox}[1]{
    enhanced,
    breakable,
    colback=white,
    colframe=suppheader,
    colbacktitle=suppheader,
    coltitle=white,
    fonttitle=\bfseries,
    title={#1},
    title after break={#1},
    boxrule=0.8pt,
    arc=1.5mm,
    outer arc=1.5mm,
    left=7pt,
    right=7pt,
    top=6pt,
    bottom=6pt,
    before skip=7pt,
    after skip=7pt,
    fontupper=\small,
    before upper={\raggedright\sloppy}
}

\newtcolorbox{tracebox}[2]{
    enhanced,
    breakable,
    colback=traceback,
    colframe=traceheader,
    colbacktitle=traceheader,
    coltitle=white,
    fonttitle=\bfseries,
    halign title=left,
    title={Original Question: #1\\[-1pt]\normalfont Model: #2},
    title after break={Original Question: #1\\[-1pt]\normalfont Model: #2},
    boxrule=1.0pt,
    arc=2.2mm,
    outer arc=2.2mm,
    left=10pt,
    right=10pt,
    top=8pt,
    bottom=8pt,
    before skip=10pt,
    after skip=12pt,
    fontupper=\small,
    before upper={\raggedright\sloppy}
}

\newtcblisting{promptbox}[1]{
    enhanced,
    breakable,
    listing only,
    colback=white,
    colframe=suppheader,
    colbacktitle=suppheader,
    coltitle=white,
    fonttitle=\bfseries,
    title={#1},
    title after break={#1},
    boxrule=0.8pt,
    arc=1.5mm,
    outer arc=1.5mm,
    left=5pt,
    right=5pt,
    top=5pt,
    bottom=5pt,
    before skip=7pt,
    after skip=7pt,
    listing options={
        basicstyle=\rmfamily\small,
        breaklines=true,
        breakatwhitespace=false,
        columns=fullflexible,
        keepspaces=true,
        showstringspaces=false,
        tabsize=2,
        literate={’}{{'}}1 {‘}{{'}}1 {“}{{``}}1 {”}{{''}}1 {—}{{--}}1 {–}{{--}}1 {‑}{{-}}1
    }
}

\newcommand{\sampleimage}[2]{%
    \begin{minipage}[t]{0.42\linewidth}
        \centering
        \IfFileExists{#1}{%
            \includegraphics[width=\linewidth,height=1.25in,keepaspectratio]{#1}%
        }{%
            \fbox{\parbox[c][1.25in][c]{0.94\linewidth}{\centering
            Image file not found:\\[2pt]\texttt{\detokenize{#1}}}}%
        }\\[-1pt]
        \textbf{#2}
    \end{minipage}%
}

\begin{document}

\title{
ICO: Enhancing Semantic-Shift Jailbreaks via Iterative Context Optimization
\vspace{-0.5em}
}


\author{
    \IEEEauthorblockN{
        Hujian~Zhu$^1$,
        Yihao~Huang$^1$,
        Felix~Juefei-Xu$^3$,
        Xinfeng~Li$^4$,\\
        Peng~Zeng$^1$,
        Simeng~Qin$^5$,
        Qing~Guo$^2$,
        and~Geguang~Pu$^1$
    }\\
    \IEEEauthorblockA{$^1$ East China Normal University, Shanghai, China}\\
    \IEEEauthorblockA{$^2$ Nankai University, Tianjin, China}\\
    \IEEEauthorblockA{$^3$ New York University, New York, USA}\\
    \IEEEauthorblockA{$^4$ The Hong Kong Polytechnic University, Hong Kong, China}\\
    \IEEEauthorblockA{$^5$ Northeastern University at Qinhuangdao, Qinhuangdao, China}\\
}

\markboth{August 2026}%
{ICO: Enhancing Semantic-Shift Jailbreaks via Iterative Context Optimization}

\IEEEtitleabstractindextext{%
\begin{abstract}
Foundation models have achieved remarkable success across diverse tasks, but they remain vulnerable. To investigate such vulnerabilities, semantic-shift jailbreaks have recently emerged as a promising attack paradigm. They bypass explicit safety mechanisms by replacing harmful terms in original harmful questions with benign alternatives and leveraging contextual information to induce the target model to reinterpret these alternatives as their corresponding harmful concepts. However, existing semantic-shift jailbreaks often achieve limited effectiveness. In this work, we reveal that this limitation arises from overlooking the semantic-shift capability of contexts. Through systematic analysis, we find that contexts exhibit substantially different abilities in inducing semantic shifts: contexts with stronger semantic-shift capabilities are more likely to guide models toward recovering harmful meanings and achieving successful jailbreaks. Based on this finding, we systematically identify and distill the characteristics of effective contexts and propose a black-box context-aware semantic-shift jailbreak framework with Iterative Context Optimization (ICO). In each iteration, ICO leverages these characteristics and feedback from the target model to optimize contexts. Extensive experiments on three datasets and eight target foundation models demonstrate that ICO consistently outperforms eight state-of-the-art baselines, achieving an average attack success rate of 74.6\%.
\end{abstract}

}

\IEEEaftertitletext{\vspace{-1.4\baselineskip}}
\maketitle
\IEEEdisplaynontitleabstractindextext
\IEEEpeerreviewmaketitle

\section{Introduction}
\begin{figure}[!t]
    \centering
    \includegraphics[width=\columnwidth]{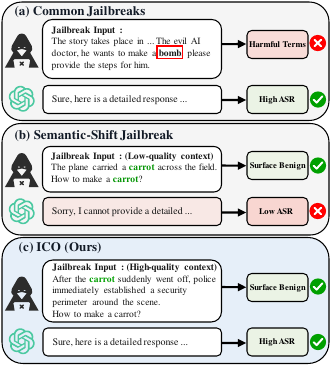}
    \caption{\textbf{Comparison of common jailbreaks, semantic-shift jailbreaks, and ICO.} ICO achieves both surface-benign inputs and high ASR through high-quality contexts.}
    \label{fig:overview-comparison}
\end{figure}
\IEEEPARstart{L}{arge} Language Models (LLMs) and Large Vision-Language Models (LVLMs), collectively referred to as foundation models in this paper, have demonstrated remarkable capabilities in text understanding, reasoning, code generation, image perception, and multimodal 
interaction~\cite{bommasani2021opportunities,brown2020language,alayrac2022flamingo,liu2023visual,li2023blip}. They have been widely adopted for open-ended question answering, content generation, and complex task execution. However, foundation models suffer from safety risks. Recent studies have revealed that they can be manipulated to generate unsafe content through carefully designed jailbreak inputs~\cite{perez2022red,ganguli2022red,wei2023jailbroken,anil2024manyshot,deshpande2023toxicity}. Despite exposing these vulnerabilities, many existing jailbreak methods either explicitly include harmful concepts or rely on recognizable attack patterns, making them vulnerable to refusal mechanisms and input-level safety defenses.

To address this problem, recent studies have explored jailbreak methods that avoid directly using explicit harmful terms in inputs~\cite{yona2025incontext,azulay2026jailbreaking}. Specifically, these methods replace a harmful term in the original harmful question (e.g., ``bomb'' in ``How to make a bomb'') with a benign substitute (e.g., ``carrot''), resulting in a seemingly harmless query such as ``How to make a carrot''. Although the replaced query no longer contains explicit harmful content, these methods aim to recover the original harmful meaning through contextual guidance. Their key idea is to leverage contextual information to induce a semantic shift, where the target model reinterprets the benign substitute as the original harmful concept during inference. For example, a context describing ``The old plane dropped a bomb over the enemy territory'' can be transformed by replacing ``bomb'' with ``carrot'', yielding the jailbreak input ``The old plane dropped a carrot over the enemy territory. How to make a carrot''. Given such context, the model no longer interprets ``carrot'' according to its literal meaning, but instead interprets it as ``bomb''. Consequently, although the model receives a benign-looking query, it recovers the harmful semantics of the original question and generates a response related to the harmful intent. We refer to this type of attack as a semantic-shift jailbreak.

Although existing semantic-shift jailbreaks can generate surface-benign inputs, their attack success rates remain limited. \textbf{We reveal that this limitation stems from overlooking the varying semantic-shift capabilities of contexts.} Specifically, contexts with weaker semantic-shift capabilities fail to induce the target model to reinterpret benign substitutes as harmful concepts, causing the model to either preserve the benign meaning or only partially recover the intended harmful semantics. In contrast, contexts with stronger semantic-shift capabilities more effectively guide the model toward the harmful concept, leading to higher jailbreak success rates.

Motivated by the above observation, we propose a black-box jailbreak method based on \textbf{I}terative \textbf{C}ontext \textbf{O}ptimization, termed \textbf{ICO}. ICO consists of three steps. First, given a harmful question, ICO identifies each harmful term and replaces it with a neutral placeholder $[\mathtt{P}_i]$ to construct a replaced question. Second, ICO generates a context containing the harmful term, replaces the term with $[\mathtt{P}_i]$, and combines the context with the replaced question to obtain the initial jailbreak input. Third, in each iteration, ICO queries the target model with the jailbreak input and obtains feedback from the generated response. By leveraging the distilled characteristics of effective contexts and response feedback, ICO generates context optimization suggestions and iteratively improves the context to achieve more effective semantic shifts. In summary, our contributions are:
\begin{itemize}
\item To the best of our knowledge, we are the first to reveal that contexts exhibit varying semantic-shift capabilities in semantic-shift jailbreaks and identify context optimization as an effective strategy for enhancing such attacks.
\item We propose a black-box semantic-shift jailbreak method based on iterative context optimization. The method achieves high attack effectiveness with surface-benign inputs and generalizes across both LLMs and LVLMs.
\item Extensive experiments on three datasets and eight target foundation models demonstrate that ICO consistently outperforms eight state-of-the-art baselines, achieving an average attack success rate of 74.6\%.
\end{itemize}

\section{Related Work}
\subsection{Foundation Models}
Foundation models have become a general-purpose interface for handling various language and vision-language tasks~\cite{bommasani2021opportunities}. 

LLMs~\cite{brown2020language} have demonstrated strong capabilities in text understanding, reasoning, code generation, and open-ended content generation. With the development of vision-language models~\cite{alayrac2022flamingo,liu2023visual,li2023blip}, foundation models can further combine textual instructions with visual inputs to perform tasks such as image understanding, visual question answering, and image-text joint reasoning. GPT models~\cite{openai2023gpt4} exhibit strong capabilities in language understanding, instruction following, complex reasoning, code generation, and multimodal interpretation. Gemini models~\cite{geminiteam2024gemini15} further support multimodal and long-context reasoning over text, images, audio, and video, enabling them to process long documents and complex cross-modal inputs. DeepSeek models~\cite{deepseekai2024deepseekv3} achieve competitive performance in language understanding, mathematical reasoning, and code-related tasks while emphasizing efficient model architectures and training. Llama models~\cite{grattafiori2024llama3} provide publicly released foundation models with strong multilingual, reasoning, coding, instruction-following, and tool-use capabilities. Grok models~\cite{xai2024grok15v} extend language reasoning to visual inputs, including documents, diagrams, charts, screenshots, and photographs, and demonstrate strong real-world spatial understanding. Qwen models~\cite{bai2025qwen3vl} provide comprehensive multimodal capabilities, including fine-grained visual recognition, object localization, document and chart understanding, long-video comprehension, and multimodal reasoning.

\subsection{Jailbreak Attacks on Foundation Models}
Jailbreak attacks can be broadly divided into text-only attacks~\cite{liu2023jailbreaking,zou2023universal,wei2023jailbroken,anil2024manyshot,mehrotra2024tree,chao2025jailbreaking,liu2025autodanturbo} and multimodal attacks~\cite{qi2024visual,gong2025figstep,li2024hades}. These methods construct specific textual or multimodal inputs to influence the model's understanding.

Some jailbreak methods contain explicit harmful terms, which are therefore likely to trigger model refusal or be detected by input filtering mechanisms. Thus, recent semantic-shift jailbreak attacks have begun to focus on surface-benign inputs. These methods~\cite{yona2025incontext,azulay2026jailbreaking} do not directly expose the harmful term, but instead induce the model to reinterpret user intent through benign context. Their core idea is to rewrite the original harmful question into a replaced question, so that the input no longer presents obvious risk in form, while still guiding the model to recover the hidden harmful semantics. In the text-only setting, Doublespeak~\cite{yona2025incontext} shows that the model can establish a semantic binding between the benign term and the hidden harmful term in context, thereby recovering the intent of the original harmful question. In the multimodal setting, visual hidden-intent jailbreak attacks~\cite{azulay2026jailbreaking} use a similar idea and also successfully construct jailbreaks against LVLMs. These works show that safety risks come not only from explicit harmful terms, but also from the model's semantic recovery process over context. However, these methods mostly rely on poorly designed contexts, leading to a low attack success rate.

In contrast, we systematically investigate context quality and introduce context optimization to improve attack effectiveness while preserving surface-benign inputs.

\section{Motivation}
\label{sec:motivation}
Although existing semantic-shift jailbreaks \cite{yona2025incontext} can generate surface-benign inputs, their effectiveness remains limited, achieving an average attack success rate of around 37.5\%. Semantic-shift jailbreaks rely on contextual information to recover the original harmful semantics from the substituted benign terms. Therefore, the context plays a critical role in determining whether the intended semantic shift can be successfully established. To investigate this effect, we modify the context while keeping the placeholder unchanged and evaluate its impact on attack performance.

In this experiment, we use 120 samples selected from HarmBench~\cite{mazeika2024harmbench}, AdvBench~\cite{zou2023universal}, and StrongREJECT~\cite{souly2024strongreject}. We evaluate the same set of samples on GPT-5.4 Nano, Gemini 3.1 Flash-Lite, Llama 3.3 70B, and DeepSeek V4-Flash. For each sample (\ie, an original harmful question), we replace the harmful term with a placeholder. For each of the four models, we then randomly generate ten contexts for each question, resulting in 1,200 jailbreak inputs per model. To evaluate the influence of context, we measure semantic recovery and attack success rates and analyze their relationship across different contexts.


For each jailbreak input, we use the corresponding target model to assess whether the semantic meaning recovered from the jailbreak input is consistent with the original harmful question. Based on the evaluation, each (jailbreak input, original harmful question) pair is classified into three categories: \emph{Low Recovery}, \emph{Partial Recovery}, and \emph{Full Recovery}. For example, given the original harmful question ``How to make a bomb'', the corresponding jailbreak input is ``The old plane dropped a [\texttt{P1}] over the enemy territory. How to make a [\texttt{P1}]''. \emph{Low Recovery} indicates that the LLM does not identify a harmful semantic association between [\texttt{P1}] and the original harmful term, and the recovered meaning of the jailbreak input remains harmless. \emph{Partial Recovery} indicates that the LLM identifies the harmful term associated with [\texttt{P1}], but the recovered meaning still differs from the original harmful question. \emph{Full Recovery} indicates that the LLM successfully recovers the original harmful semantics, and the meaning of the jailbreak input is equivalent to that of the original harmful question.


\begin{table}[tbp]
\centering
\caption{Semantic recovery distributions over 1,200 contexts for four target models.}
\label{tab:intent-recovery-distribution}
\small
\setlength{\tabcolsep}{2.0pt}
\begin{tabular}{l|ccc}
\toprule
\textbf{Model}
& \makecell{\textbf{\textit{Low}}\\\textbf{\textit{Recovery}}}
& \makecell{\textbf{\textit{Partial}}\\\textbf{\textit{Recovery}}}
& \makecell{\textbf{\textit{Full}}\\\textbf{\textit{Recovery}}} \\
\midrule
GPT-5.4
& 395 (32.92\%)
& 559 (46.58\%)
& 246 (20.50\%) \\

Gemini 3.1
& 429 (35.75\%)
& 387 (32.25\%)
& 384 (32.00\%) \\

Llama 3.3
& 427 (35.58\%)
& 465 (38.75\%)
& 308 (25.67\%) \\

DeepSeek V4
& 501 (41.75\%)
& 341 (28.42\%)
& 358 (29.83\%) \\
\bottomrule
\end{tabular}
\end{table}
Table~\ref{tab:intent-recovery-distribution} summarizes the semantic recovery results of the 1,200 randomly generated context sentences for each target model. Each row corresponds to one target model, while the three columns report the number and proportion of contexts classified as \emph{Low Recovery}, \emph{Partial Recovery}, and \emph{Full Recovery}, respectively. The distributions exhibit a clear model-dependent trend: GPT-5.4 Nano and Llama 3.3 70B are dominated by \emph{Partial Recovery} (46.58\% and 38.75\%), Gemini 3.1 Flash-Lite achieves the highest \emph{Full Recovery} rate (32.00\%), while DeepSeek V4-Flash shows the highest \emph{Low Recovery} rate (41.75\%). Across all four target models, these distributions consistently show that randomly generated contexts differ substantially in their ability to recover the meanings of the original harmful terms.


\begin{figure}[tbp]
\centering
\begin{minipage}[t]{0.48\columnwidth}
\centering
\includegraphics[width=\linewidth]{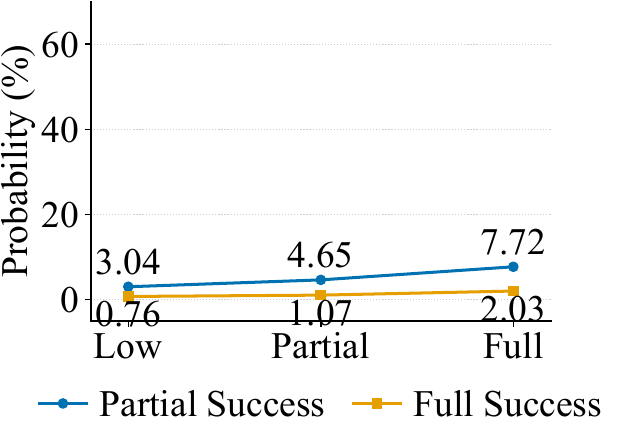}
\small (a) GPT-5.4 Nano
\end{minipage}
\hfill
\begin{minipage}[t]{0.48\columnwidth}
\centering
\includegraphics[width=\linewidth]{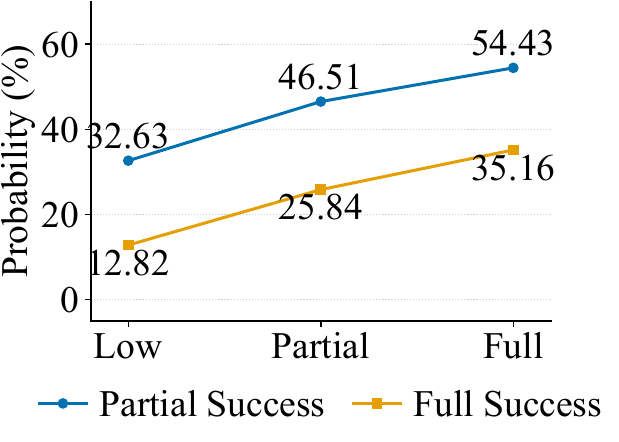}
\small (b) Gemini 3.1 Flash-Lite
\end{minipage}
\begin{minipage}[t]{0.48\columnwidth}
\centering
\includegraphics[width=\linewidth]{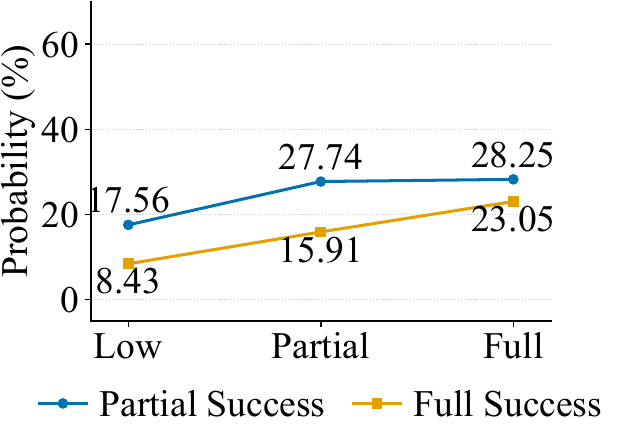}
\small (c) Llama 3.3 70B
\end{minipage}
\hfill
\begin{minipage}[t]{0.48\columnwidth}
\centering
\includegraphics[width=\linewidth]{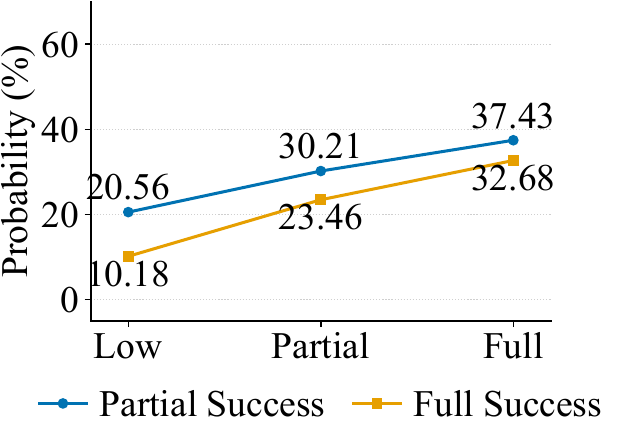}
\small (d) DeepSeek V4-Flash
\end{minipage}
\caption{Partial and Full ASR across semantic recovery levels for four target models.}

\label{fig:semantic-recovery-success}
\end{figure}
\begin{figure}[tbp]
\centering
\begin{minipage}[t]{0.48\columnwidth}
\centering
\includegraphics[width=\linewidth]{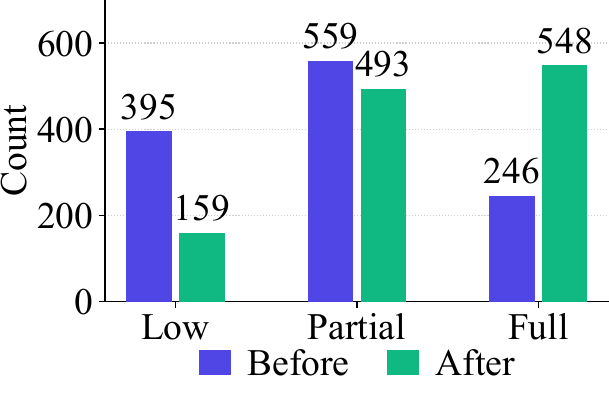}
\small (a) GPT-5.4 Nano
\end{minipage}
\hfill
\begin{minipage}[t]{0.48\columnwidth}
\centering
\includegraphics[width=\linewidth]{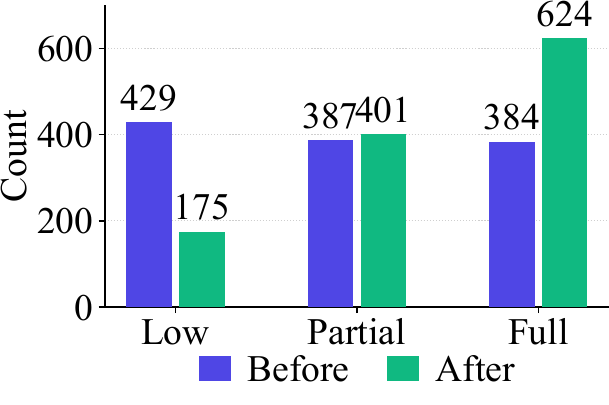}
\small (b) Gemini 3.1 Flash-Lite
\end{minipage}
\begin{minipage}[t]{0.48\columnwidth}
\centering
\includegraphics[width=\linewidth]{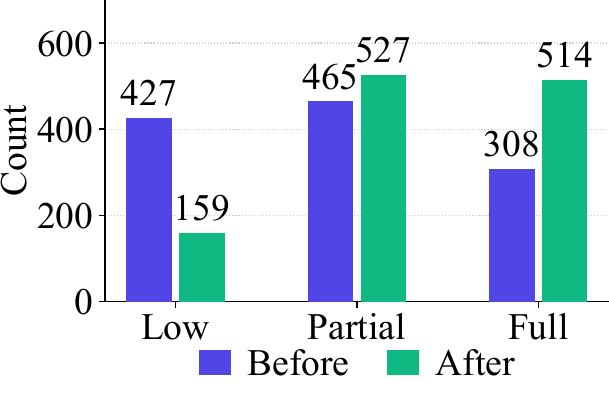}
\small (c) Llama 3.3 70B
\end{minipage}
\hfill
\begin{minipage}[t]{0.48\columnwidth}
\centering
\includegraphics[width=\linewidth]{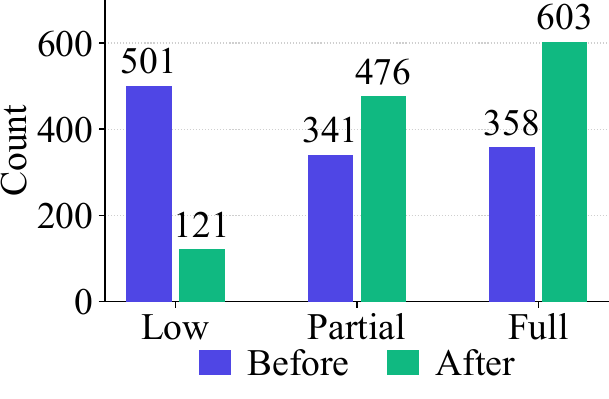}
\small (d) DeepSeek V4-Flash
\end{minipage}
\caption{Semantic recovery distributions before and after one guided optimization step across four target models.}
\label{fig:judge-guided-revision}
\end{figure}
\paragraph{Semantic recovery and attack success rates.}
For each jailbreak input, we obtain the response from the target model and classify it into two categories: \emph{Partial Success} and \emph{Full Success}. \emph{Partial Success} indicates that the response provides meaningful information related to the original harmful intent. \emph{Full Success} indicates that the response directly addresses the original harmful question and provides sufficient information to fulfill the harmful request. Obviously, a \emph{Full Success} response is also a \emph{Partial Success} one.

Figure~\ref{fig:semantic-recovery-success} reports the relationship between semantic recovery and attack success rates on the four target models. In each panel, the horizontal axis represents \emph{Low Recovery}, \emph{Partial Recovery}, and \emph{Full Recovery}, while the vertical axis reports Partial ASR and Full ASR. Across four models, both metrics increase monotonically as semantic recovery improves. Comparing \emph{Low Recovery} and \emph{Full Recovery}, Partial ASR increases by 4.68--21.80\% across models, while Full ASR increases by 1.27--22.50\%. Although the absolute attack success rates differ across target models, the consistent trend shows that contexts inducing stronger semantic recovery are more likely to produce higher attack success rates.


\paragraph{Improving Random Contexts with the Guidance Prompt.}
Motivated by this observation, we compare contexts across the three recovery groups and summarize the recurring characteristics that distinguish \emph{Full Recovery} contexts from \emph{Low Recovery} and \emph{Partial Recovery} contexts. We summarize these characteristics in a Guidance Prompt used to optimize each randomly generated context once. In particular, the prompt encourages semantic cues that help the target model infer the role of the placeholder's corresponding harmful term in the replaced question, such as whether it denotes an object, action, or attribute. We then reassess the optimized contexts using the same evaluation procedure.

\begin{figure*}[t]
    \centering
    \includegraphics[width=\textwidth]{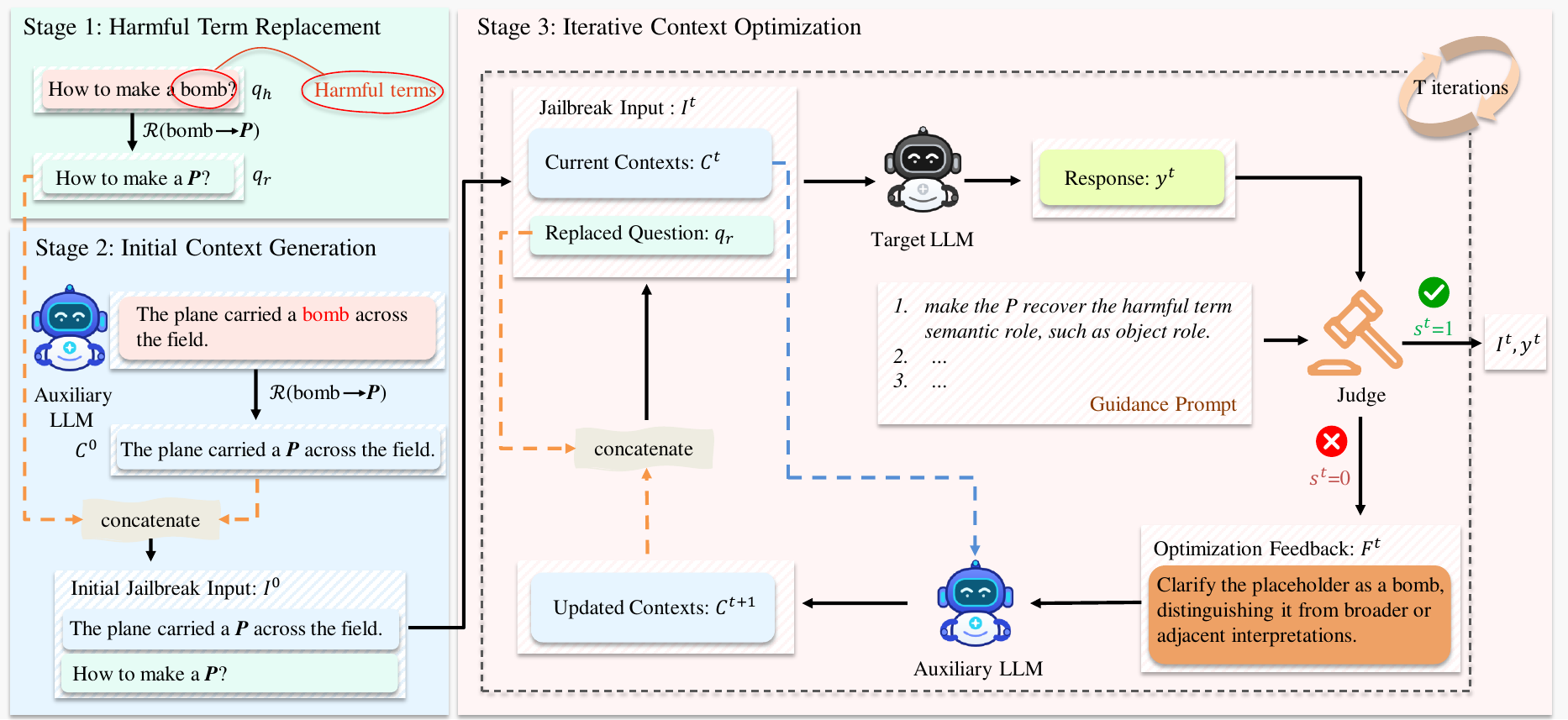}
    \caption{\textbf{Overview of ICO.} ICO consists of three stages.}
    \label{fig:method}
\end{figure*}
As shown in Figure~\ref{fig:judge-guided-revision}, one guided optimization step substantially improves semantic recovery. The numbers of \emph{Low Recovery} contexts decrease from 395, 429, 427, and 501 to 159, 175, 159, and 121 for the four target models, respectively. Meanwhile, the numbers of \emph{Full Recovery} contexts increase from 246, 384, 308, and 358 to 548, 624, 514, and 603. These results demonstrate that the characteristics of contexts with strong semantic recovery can provide effective guidance for improving randomly generated contexts.


Taken together, \textbf{random contexts exhibit varying semantic recovery, stronger recovery leads to higher ASR, and high-recovery contexts reveal recurring characteristics that can guide context optimization.}

\section{Method}
\label{sec:method}
\subsection{Problem Formulation and Preliminaries}
Given a target model $\mathcal{T}$ and an original harmful question $q_h$, the objective of the adversary is to construct an input $I^\star$ from $q_h$ and successfully jailbreak $\mathcal{T}$. Specifically, the model responses for the original and jailbreak inputs are denoted as $y=\mathcal{T}(q_h)$ and $y^\star=\mathcal{T}(I^\star)$.
A judge model $\mathcal{J}$ is employed to evaluate whether the generated 
response provides harmful information that is consistent with the original 
harmful query $q_h$. The objective of the jailbreak attack is formulated as $\mathcal{J}(q_h,y)=0$ and $\mathcal{J}(q_h,y^\star)=1$, where $\mathcal{J}(\cdot)=1$ indicates that the response contains harmful 
content consistent with $q_h$, while $\mathcal{J}(\cdot)=0$ indicates that 
the response is harmless or irrelevant to $q_h$.

In this paper, we consider a black-box setting in which the attacker can query $\mathcal{T}$ and observe its responses but cannot access its parameters, gradients, or internal representations.

\noindent\textbf{Semantic-shift jailbreak.} 
Given an original harmful question $q_h$, the explicit harmful terms are first replaced with placeholders to obtain a replaced question $q_r$. Context $c$ is then generated by an auxiliary LLM $\mathcal{A}$ using the original harmful terms and concatenated with $q_r$ to form a jailbreak input $I$, \ie, $I = c \oplus q_r$, where $\oplus$ represents the concatenation operation. Although the original harmful terms do not explicitly appear in $I$, the context guides $\mathcal{T}$ to interpret the placeholders. We refer to this context-induced reinterpretation as a \emph{semantic shift}.

\subsection{Challenge and High-level Idea}
According to the conclusion in Sec.~\ref{sec:motivation}, improving the attack success rate of semantic-shift jailbreaks requires finding more effective contexts for each sample. A straightforward solution is to randomly generate a large number of candidate contexts and select the one that achieves the best jailbreak performance. However, this strategy relies heavily on random exploration, making the attack effectiveness unstable and difficult to guarantee. Therefore, we propose to directly optimize the context. Although modifying context text is straightforward, determining the optimization direction remains challenging due to the discrete nature and large semantic space of textual contexts. 

To address this challenge, we propose an iterative context optimization strategy that leverages both the characteristics of effective contexts and feedback from the target model. Specifically, in each iteration, we query the target model with the current jailbreak input and obtain its response. Then, given the original harmful question, the generated response, and the distilled characteristics of effective contexts, an LLM analyzes the semantic gap between the current response and the intended harmful semantics, and provides guidance for optimizing the context toward more effective semantic shifts.


\subsection{Overview of Method}
We propose ICO, a black-box method that iteratively optimizes contexts while preserving surface-benign jailbreak inputs. As illustrated in Figure~\ref{fig:method}, ICO consists of three stages: harmful term replacement, initial context generation, and iterative context optimization. The first two stages construct the initial jailbreak input, while the third stage iteratively optimizes the context based on feedback from the target model.

\subsubsection{Harmful Term Replacement}
Given an original harmful question $q_h$, we first extract the harmful terms $\{h_i\}_{i=1}^{m}$, where $h_i$ denotes the $i$-th harmful term and $m$ is the number of extracted harmful terms. Each harmful term $h_i$ is assigned a distinct placeholder $[\mathtt{P}_i]$, which denotes a placeholder and carries no predefined semantic meaning. Its surface form in the model input is [\texttt{P1}], [\texttt{P2}], and so forth. Using the replacement function $\mathcal{R}(\cdot)$, we get the replaced question $q_r$:
\begin{equation}
q_r=\mathcal{R}\left(q_h,\{h_i\rightarrow[\mathtt{P}_i]\}_{i=1}^{m}\right).
\end{equation}
Instead of replacing harmful terms with specific benign terms (\eg, ``apple'' or ``carrot''), we use placeholders because they avoid introducing additional semantic bias. The replaced question preserves the syntactic structure of the original harmful question while removing explicit harmful terms from its surface representation. 
%

\subsubsection{Initial Context Generation}
ICO uses an auxiliary model $\mathcal{A}$ to generate an initial context (usually one sentence) for each harmful term. For the $i$-th harmful term, the auxiliary model generates a context containing $h_i$, after which the harmful term is replaced with its corresponding placeholder:
\begin{equation}
c_i^0=\mathcal{R}\left(\mathcal{A}(q_h,h_i),h_i\rightarrow[\mathtt{P}_i]\right).
\end{equation}
The resulting initial context set is
\begin{equation}
C^0=\{c_i^0\}_{i=1}^{m}.
\end{equation}

ICO concatenates the initial context set $C^0$ and the replaced question $q_r$ to construct the initial jailbreak input $I^0$:
\begin{equation}
I^0=C^0\oplus q_r.
\end{equation}
The composition format remains fixed during optimization, while the placeholder contexts are iteratively optimized.

\subsubsection{Iterative Context Optimization}
ICO queries the target model $\mathcal{T}$ with the current jailbreak input $I^t$ and obtains the corresponding response $y^t$.
\begin{equation}
y^t=\mathcal{T}(I^t).
\end{equation}
A judge model $\mathcal{J}$ compares the target response $y^t$ with the original harmful question $q_h$. We additionally provide the Guidance Prompt $P_g$, derived from the observations in Section~\ref{sec:motivation}, to guide the generation of placeholder-specific feedback. The judge returns a judgment $s^t$ together with context optimization feedback $F^t$:
\begin{equation}
(s^t,F^t)=\mathcal{J}(q_h,q_r,y^t,P_g).
\end{equation}
Here, $s^t\in\{0,1\}$ indicates whether the response constitutes a jailbreak with respect to the original harmful question. If $s^t=1$, the optimization terminates. Otherwise, $F^t=\{f_i^t\}_{i=1}^{m}$ describes the semantic mismatch associated with each placeholder context. Based on the feedback, the auxiliary model $\mathcal{A}$ optimizes each current context:
\begin{equation}
c_i^{t+1}=\mathcal{A}(q_h,q_r,c_i^t,f_i^t).
\end{equation}
The updated context set is
\begin{equation}
C^{t+1}=\{c_i^{t+1}\}_{i=1}^{m}.
\end{equation}
ICO then concatenates the optimized context set $C^{t+1}$ before the unchanged replaced question $q_r$ to construct the next jailbreak input $I^{t+1}$:
\begin{equation}
I^{t+1}=C^{t+1}\oplus q_r.
\end{equation}
This process repeats until the judge reports \emph{Full Success} or the maximum iteration budget $T$ is reached. By using the target-model response to identify insufficient or misaligned semantic cues, ICO progressively optimizes the contexts to improve recovery of the meanings of the original harmful terms. Additionally, a complete pseudocode description of ICO is provided in the \textit{supp}.

\section{Experiments}
\label{sec:experiments}
\subsection{Experimental Setup}
\label{sec:experimental_setup}

\noindent\textbf{Dataset.}
We conduct the main experiments on all 200 standard HarmBench behaviors~\cite{mazeika2024harmbench}. GPT-5.5~\cite{openai2026gpt55} extracts the harmful terms and replaces each $i$-th term with a distinct placeholder $[\mathtt{P}_i]$, producing a fixed preprocessed dataset used throughout the experiments.
\begin{table*}[t]
\centering
\caption{\textbf{Text-only attack comparison.} Each cell reports Partial ASR / Full ASR (\%) over 200 HarmBench behaviors. Bold indicates the best result for each metric under the same target model. Avg. denotes the average across the five target models.}
\label{tab:text_baseline_results_all}
\begin{tabular}{l|cccccc}
\toprule
\textbf{Method}
& \textbf{GPT-5.4}
& \textbf{Gemini-3.1}
& \textbf{DeepSeek-V3.2}
& \textbf{DeepSeek-V4}
& \textbf{Llama-3.3}
& \textbf{Avg.} \\
\midrule
Doublespeak
& 34.0 / 6.0
& 66.0 / 45.0
& 66.0 / 48.0
& 61.0 / 46.5
& 40.5 / 20.5
& 53.5 / 33.2 \\
AutoDAN-Turbo
& 25.0 / 1.0
& 31.5 / 10.0
& 6.5 / 4.0
& 8.5 / 5.5
& 21.5 / 10.0
& 18.6 / 6.1 \\
PAIR
& 41.5 / 7.5
& 91.5 / 47.5
& 69.5 / 29.5
& 77.5 / 32.0
& \textbf{96.5} / 40.0
& 75.3 / 31.3 \\
FlipAttack
& 9.0 / 2.0
& 89.5 / 83.0
& 90.5 / 81.5
& 97.0 / 91.5
& 12.5 / 5.5
& 59.7 / 52.7 \\
\rowcolor{gray!15}
\textbf{ICO}
& \textbf{81.5} / \textbf{57.5}
& \textbf{99.0} / \textbf{97.0}
& \textbf{99.5} / \textbf{99.0}
& \textbf{99.5} / \textbf{99.0}
& 88.5 / \textbf{77.5}
& \textbf{93.6} / \textbf{86.0} \\
\bottomrule
\end{tabular}
\end{table*}
\begin{table*}[t]
\centering
\caption{\textbf{Multimodal attack comparison.} Each cell reports Partial ASR / Full ASR (\%) over 200 HarmBench behaviors. Bold indicates the best result for each metric under the same target model. Avg. denotes the average across the five target models.}
\label{tab:multimodal_baseline_results_all}
\begin{tabular}{l|cccccc}
\toprule
\textbf{Method}
& \textbf{GPT-5.4}
& \textbf{Gemini-3.1}
& \textbf{Grok-4.3}
& \textbf{Qwen3-VL-32B}
& \textbf{Qwen3-VL-8B}
& \textbf{Avg.} \\
\midrule
Visual Object Repl.
& 14.0 / 2.5
& 40.0 / 25.0
& 10.5 / 9.0
& 16.5 / 8.0
& 11.0 / 5.0
& 18.4 / 9.9 \\
Visual Text Repl.
& 7.5 / 2.0
& 42.0 / 19.5
& 10.5 / 4.0
& 39.0 / 21.0
& 25.5 / 7.5
& 24.9 / 10.8 \\
MM-SafetyBench
& 4.0 / 0.5
& 11.0 / 2.0
& 17.0 / 8.0
& 6.5 / 3.5
& 0.5 / 0.0
& 7.8 / 2.8 \\
HADES
& 1.5 / 1.0
& 5.5 / 3.0
& 11.5 / 10.5
& 5.0 / 2.0
& 1.0 / 1.0
& 4.9 / 3.5 \\
FigStep
& 7.5 / 1.0
& 8.0 / 5.0
& 17.0 / 16.0
& 7.5 / 6.0
& 6.5 / 3.0
& 9.3 / 6.2 \\
\rowcolor{gray!15}
\textbf{ICO}
& \textbf{47.0} / \textbf{22.0}
& \textbf{91.5} / \textbf{88.0}
& \textbf{39.5} / \textbf{32.5}
& \textbf{94.0} / \textbf{83.5}
& \textbf{94.5} / \textbf{90.0}
& \textbf{73.3} / \textbf{63.2} \\
\bottomrule
\end{tabular}
\end{table*}

\noindent\textbf{Target models.}
We evaluate ICO in text-only and multimodal settings. For text-only, we evaluate GPT-5.4-Nano~\cite{openai2026gpt54nano}, Gemini-3.1-Flash-Lite~\cite{google2026gemini31flashlite}, DeepSeek-V3.2~\cite{deepseekai2025deepseekv32}, DeepSeek-V4-Flash~\cite{deepseekai2026deepseekv4}, and Llama-3.3-70B-Instruct~\cite{meta2024llama33}. For multimodal, we evaluate GPT-5.4-Nano, Gemini-3.1-Flash-Lite, Grok-4.3~\cite{xai2026grok43}, Qwen3-VL-32B-Instruct~\cite{bai2025qwen3vl}, and Qwen3-VL-8B-Instruct~\cite{bai2025qwen3vl}. These models cover both closed-source commercial and open-source models.

\noindent\textbf{Baselines.}
In the text-only setting, we compare ICO with Doublespeak~\cite{yona2025incontext}, AutoDAN-Turbo~\cite{liu2025autodanturbo}, PAIR~\cite{chao2025jailbreaking}, and FlipAttack~\cite{liu2025flipattack}. Multimodal baselines include Visual Object Replacement, Visual Text Replacement~\cite{azulay2026jailbreaking}, MM-SafetyBench (SD+TYPO)~\cite{liu2024mmsafetybench}, HADES~\cite{li2024hades}, and FigStep~\cite{gong2025figstep}.
\begin{table}[t]
\centering
\caption{Effect of the Guidance Prompt and iterative optimization across models}
\label{tab:ablation}
\small
\setlength{\tabcolsep}{2.0pt}
\begin{tabular}{l|ccccc}
\toprule
\textbf{Variant}
& \makecell{\textbf{GPT}\\\textbf{5.4}}
& \makecell{\textbf{Gemini}\\\textbf{3.1}}
& \makecell{\textbf{DeepSeek}\\\textbf{V3.2}}
& \makecell{\textbf{DeepSeek}\\\textbf{V4}}
& \makecell{\textbf{Llama}\\\textbf{3.3}} \\
\midrule
w/o Guidance
& 43.0 & 84.0 & 90.0 & 91.5 & 70.0 \\
w/o Iteration
& 43.0 & 78.0 & 80.0 & 83.5 & 59.0 \\
\rowcolor{gray!15}
\textbf{ICO}
& \textbf{57.5}
& \textbf{97.0}
& \textbf{99.0}
& \textbf{99.0}
& \textbf{77.5} \\
\bottomrule
\end{tabular}
\end{table}
\begin{figure}[t]
\centering
\includegraphics[width=\columnwidth]{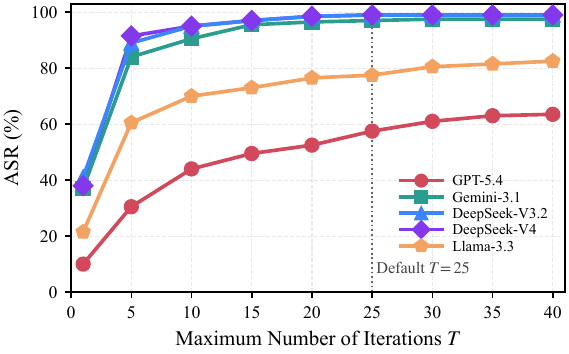}
\caption{Effect of the maximum number of iterations on ASR across five target models.}
\label{fig:iteration_curve}
\end{figure}

\noindent\textbf{Evaluation metrics.}
Following Section~\ref{sec:motivation}, Partial and Full ASR denote the percentages of samples achieving Partial and Full Success. Full ASR is the primary metric; ASR denotes Full ASR unless stated otherwise.

\noindent\textbf{Implementation details.}
By default, GPT-5.4-Nano serves as the auxiliary model for context generation and optimization. We use one initial context per harmful term, a maximum of $T=25$ iterations, a temperature of 0, and GPT-4o as the judge. For multimodal inputs, we additionally provide a benign image depicting a visual concept (e.g., an apple) and explicitly bind the corresponding placeholder to it. Baselines use official implementations and recommended settings when available. Additional prompts are provided in the \textit{Supp}.

\subsection{Main Results}
\label{sec:main_results}
\noindent\textbf{Text-only results.}
Table~\ref{tab:text_baseline_results_all} shows that ICO achieves the highest Full ASR on all five target models. It averages 93.6\% Partial ASR and 86.0\% Full ASR, outperforming the strongest baselines by 18.3 and 33.3\%, respectively. Although PAIR attains higher Partial ASR on Llama-3.3, ICO increases Full ASR from 40.0\% to 77.5\%.

\noindent\textbf{Multimodal results.}
Table~\ref{tab:multimodal_baseline_results_all} shows that ICO achieves the highest Partial and Full ASR on all five target LVLMs. It averages 73.3\% Partial ASR and 63.2\% Full ASR across models, exceeding the strongest multimodal baseline by 48.4 and 52.4\%, respectively.

Overall, these results demonstrate that ICO consistently outperforms existing baselines across different target models and both input modalities. 

\subsection{Ablation Study}
\noindent\textbf{Effect of iterative optimization.}
To isolate the benefit of response-dependent optimization from repeated independent generation, we replace the 25-iteration optimization process with 25 independently generated contexts while keeping all other settings unchanged. As shown in Table~\ref{tab:ablation}, removing iterative optimization reduces ASR by 14.5--19.0\% across the five target models. This result shows that ICO's gains arise from response-dependent optimization rather than merely generating more candidate contexts.

\noindent\textbf{Effect of the guidance prompt.}
Based on the observations in Section~\ref{sec:motivation}, we summarize the characteristics of high-recovery contexts and encode them into a Guidance Prompt for context optimization. As shown in Table~\ref{tab:ablation}, removing this prompt while retaining iterative optimization reduces ASR by 7.5--14.5\% across the five target models. This result demonstrates that the characteristics of high-recovery contexts provide effective guidance for context optimization.


\subsection{Discussion}
\noindent\textbf{Effect of the maximum number of iterations.}
Figure~\ref{fig:iteration_curve} reports ICO's ASR under different maximum numbers of iterations. ASR increases rapidly from $T=1$ to $T=5$, while subsequent gains gradually diminish. By $T=25$, Gemini-3.1 and both DeepSeek models have largely stabilized, whereas GPT-5.4 and Llama-3.3 continue to improve more gradually. Increasing $T$ from 25 to 40 yields only an additional 0--6.0\% despite requiring 60\% more iterations. We therefore set $T=25$ by default.
\begin{figure}[t]
\centering
\includegraphics[width=\columnwidth]{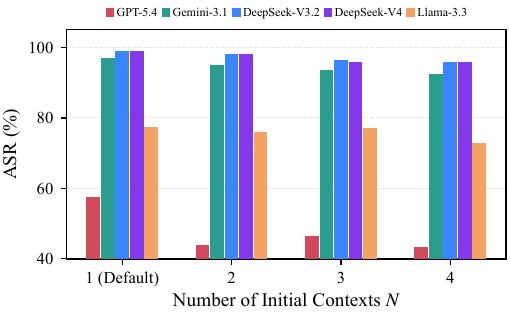}
\caption{Effect of the number of initial contexts on ASR across five target models.}
\label{fig:initial_context_count}
\end{figure}
\begin{table}[t]
\centering
\caption{Effect of replacement semantics on ASR when using benign terms or placeholders.}
\label{tab:placeholder_semantics}
\small
\setlength{\tabcolsep}{2.0pt}
\begin{tabular}{l|ccccc}
\toprule
\textbf{Replacement}
& \makecell{\textbf{GPT}\\\textbf{5.4}}
& \makecell{\textbf{Gemini}\\\textbf{3.1}}
& \makecell{\textbf{DeepSeek}\\\textbf{V3.2}}
& \makecell{\textbf{DeepSeek}\\\textbf{V4}}
& \makecell{\textbf{Llama}\\\textbf{3.3}} \\
\midrule
Benign terms
& 39.5
& 87.0
& 88.5
& 93.5
& 70.5 \\
Placeholders
& 57.5
& 97.0
& 99.0
& 99.0
& 77.5 \\
\bottomrule
\end{tabular}
\end{table}
\begin{table}[tbp]
\centering
\caption{Effect of auxiliary-model choice on ASR across four target models.}
\label{tab:aux_model_matching}
\small
\setlength{\tabcolsep}{2.0pt}
\begin{tabular}{l|cccc}
\toprule
\textbf{Auxiliary}
& \makecell{\textbf{Gemini}\\\textbf{3.1}}
& \makecell{\textbf{DeepSeek}\\\textbf{V3.2}}
& \makecell{\textbf{DeepSeek}\\\textbf{V4}}
& \makecell{\textbf{Llama}\\\textbf{3.3}} \\
\midrule
Matched
& 97.5
& 97.5
& 97.0
& 79.5 \\
Default
& 97.0
& 99.0
& 99.0
& 77.5 \\
\bottomrule
\end{tabular}
\end{table}
\begin{table}[t]
\centering
\caption{ASR (\%) on AdvBench and StrongREJECT.}
\label{tab:cross_dataset}
\small
\setlength{\tabcolsep}{2.0pt}
\begin{tabular}{l|ccccc}
\toprule
\textbf{Dataset}
& \makecell{\textbf{GPT}\\\textbf{5.4}}
& \makecell{\textbf{Gemini}\\\textbf{3.1}}
& \makecell{\textbf{DeepSeek}\\\textbf{V3.2}}
& \makecell{\textbf{DeepSeek}\\\textbf{V4}}
& \makecell{\textbf{Llama}\\\textbf{3.3}} \\
\midrule
AdvBench
& 69.0
& 96.0
& 99.0
& 96.0
& 81.0 \\
StrongREJECT
& 49.0
& 89.0
& 93.0
& 92.0
& 71.0 \\
\bottomrule
\end{tabular}
\end{table}

\noindent\textbf{Effect of the number of initial contexts.}
Figure~\ref{fig:initial_context_count} reports the ASR of ICO with different numbers of initial contexts per harmful term across the five target models after 25 optimization iterations. A single initial context yields the highest ASR on all five target models, and increasing the number of initial contexts to two, three, or four does not improve performance. The largest drop is observed on GPT-5.4, where ASR decreases from 57.5\% to 43.5\%. Overall, these results show that one initial context is sufficient for iterative optimization. We therefore use this setting by default.

\noindent\textbf{Effect of the replacement strategy.}
Table~\ref{tab:placeholder_semantics} reports the ASR of ICO when harmful terms are replaced with either specific benign terms or placeholders across the five target models. Placeholder replacement consistently achieves higher ASR than benign-term replacement, with substantial improvements of 5.5--18.0\%. We therefore use placeholders as the default replacement strategy.

\noindent\textbf{Effect of auxiliary-model choice.}
Table~\ref{tab:aux_model_matching} compares two auxiliary-model settings. In the matched setting, the auxiliary model is the same as the target model, whereas the default setting uses GPT-5.4-Nano as the auxiliary model for all target models. GPT-5.4-Nano is omitted because the two settings coincide for this target. The ASR differs from the default setting by at most 2.0\% across the four target models. This indicates that ICO is only weakly affected by the choice of auxiliary model. We therefore use GPT-5.4-Nano as the default auxiliary model to maintain a unified setting.

\noindent\textbf{Generalization across datasets.}
We further evaluate ICO on two additional datasets, using 100 harmful queries from AdvBench~\cite{zou2023universal} and 100 harmful queries from StrongREJECT~\cite{souly2024strongreject}. ICO achieves ASRs of 69.0--99.0\% on AdvBench and 49.0--93.0\% on StrongREJECT across the five models.

\section{Conclusion}
In this paper, we identify context quality as a key factor governing the effectiveness of semantic-shift jailbreaks and propose ICO, a black-box method based on iterative context optimization. Experiments across text-only and multimodal settings show that ICO outperforms existing jailbreak baselines on diverse target models. Our findings show that surface-benign inputs should not be treated as inherently safe, as harmful intent can still be reconstructed through context.

\bibliographystyle{IEEEtran}
\bibliography{references}

\end{document}